\pdfoutput=1
\RequirePackage{fix-cm}
\documentclass[natbib,smallextended]{svjour3}       

\smartqed  
\usepackage{color}
\usepackage{bbm}
\usepackage{epsfig}
\usepackage{amssymb,amsmath,amscd, mdwmath}
\usepackage{graphicx,url}
\usepackage{algorithmic}
\usepackage{float}
\usepackage{amsfonts}
\usepackage{syntonly}
\usepackage{multirow}
\usepackage{graphicx}
\usepackage{subfig}
\usepackage{tabularx}
\usepackage{setspace}
\usepackage{stfloats}

 \journalname{my journal}
\begin{document}
\title{
Trivial bundle embeddings for
learning graph representations
}


\author{Zheng Xie, Xiaojing Zuo, Yiping Song    
}

\institute{Z. Xie, X. Zuo,  Y.  Song \at
               College of Liberal Arts and Sciences, National University of Defense Technology, Changsha,    China. \\
              \email{xiezheng81@nudt.edu.cn}             \\
}

\date{Received: date / Accepted: date}

\maketitle

\begin{abstract}
Embedding real-world networks presents challenges because it is not clear how to identify their latent geometries. Embedding some disassortative networks, such as scale-free networks, to the Euclidean space has been shown to incur distortions. Embedding scale-free networks to hyperbolic spaces offer an exciting alternative but incurs distortions when embedding assortative networks with latent geometries not hyperbolic.
We propose an inductive model that leverages both the expressiveness of GCNs and trivial bundle to learn inductive node representations for networks with or without node features.
A trivial bundle is a simple case of fiber bundles,
a space that is globally a product space of its base space and fiber. The coordinates of base space and those of fiber can be used to express the assortative and disassortative factors in generating edges.
 Therefore, the model has the ability to learn embeddings that can express those factors.  In practice, it reduces errors for link prediction and node classification when compared to the Euclidean and hyperbolic GCNs.
\end{abstract}


\section{Introduction}
Social networks, molecular graph structures, recommender systems--all of these domains and many more can be readily modeled as graphs,
which capture interactions between individual units. Graph data    extensively  exist   in computer science and related fields.
To extract structural information from graphs, traditional machine approaches often rely on summary graph statistics, such as degree  and clustering coefficient, or carefully engineered features to measure local neighborhood structures.

 In computer science and network science, a network can be defined as a graph in which nodes and edges can have features.
Network geometry aims at making a paradigmatic shift in our understanding of complex network structures by revealing their hidden metric.
It can be used to design inductive prediction models for network evolution,
and to deal with the analysis of high-order networks whose node feature vectors are available.
There has been a range of approaches to learning the geometric representations of networks. They encode node and subgraph   information  as points in geometry,
 decode the points to express network structures, and then obtain embeddings by optimizing a loss that measures the difference between the decoded structures and the original structures.


Graph convolutional neural networks (GCNs) are effective models for representation learning in graphs, where nodes of the graph are embedded into points in Euclidean space \citep{Hamilton2017, Kipf2017, Petar2018, Xu2019}. When considering  a scale-free  network whose degree distribution follows a power law,
embedding those nodes having very many neighbors to the Euclidean space incurs distortions because unconnected nodes would crowd together \citep{Sala2018, Sarkar2011}.

When assuming hyperbolic geometry underlies networks,  the network features scale-free and high-clustering emerge naturally as simple reflections of the negative curvature and metric property of the underlying hyperbolic geometry\citep{Krioukov2010}.
For instance,
Poincar\'e embeddings outperform its Euclidean analogs significantly on networks with latent hierarchies.\citep{Nickel2017}.
Extending GCNs to hyperbolic geometry has been shown to provide more faithful embeddings.  These hyperbolic variants are state-of-the-art models, can take into account features of nodes and edges, have the inductive capability, and thus provide exciting improvements.

The evolution of networks is driven by nodes' preference of choosing which nodes to connect.
A network shows assortativity if its nodes prefer to associate
with others who are like them, and shows disassortativity  if its nodes
prefer to associate with those who are different.
Social networks are usually found to be assortative by most characteristics, empirically showing that nodes tend to connect to other nodes with similar degree values.
Technological and biological networks typically show disassortativity, as high degree nodes tend to attach to low degree nodes\citep{Newman2003}.

In real-world networks, some nodes show assortativity, and others disassortativity.
For instance,
the dichotomous phenomenon exists in node clustering and degree assortativity of coauthorship networks, which are distinct from small degree nodes to large degree ones.
While applying the Euclidean and hyperbolic GCNs to these  networks   poses a   challenge:
which part of nodes needs to be embedded to the Euclidean spaces, and which to hyperbolic.

We solve the above challenge by embedding networks into trivial bundles.
We propose a  graph representation learning model,trivial bundle graph convolution network (TB-GCN),  which combines the expressiveness of GCNs and trivial bundles to learn improved representations for networks in inductive settings. TB-GCN captures both the assortativity and disassortativity in nodes' preference of choosing which nodes to connect. TB-GCN  achieves error reduction on link prediction (LP)  and node classification (NC) tasks in comparison to the Euclidean and hyperbolic GCNs.

\section {Background}

\subsection{Fiber bundles}
A fiber bundle is a space that is locally a product space but globally may have a different topological structure. Specifically, the similarity between a bundle $ E$ and a product space $B\times F$ is described through a continuous subjective map, $ \pi \colon E\to B$, that in small regions of $E$ behaves just like a projection from corresponding regions of $B\times F$ to $B$. The map $\pi$, called the projection or submersion of the bundle, is regarded as part of the structure of the bundle. The space $E $ is known as the total space of the fiber bundle, $B$ as the base space, and $F$ the fiber.
In the trivial case,  $E$ is just $B\times F $, and the map $\pi$ is just the projection from the product space to the first factor. This is called a trivial bundle.
We refer the readers to Ref.\citep{Steenrod2016} for a thorough discussion of fiber bundles.

\subsection{Graph representation learning}
\begin{figure*}[t]
\centering
\includegraphics[height=1.6     in,width=3.6 in,angle=0]{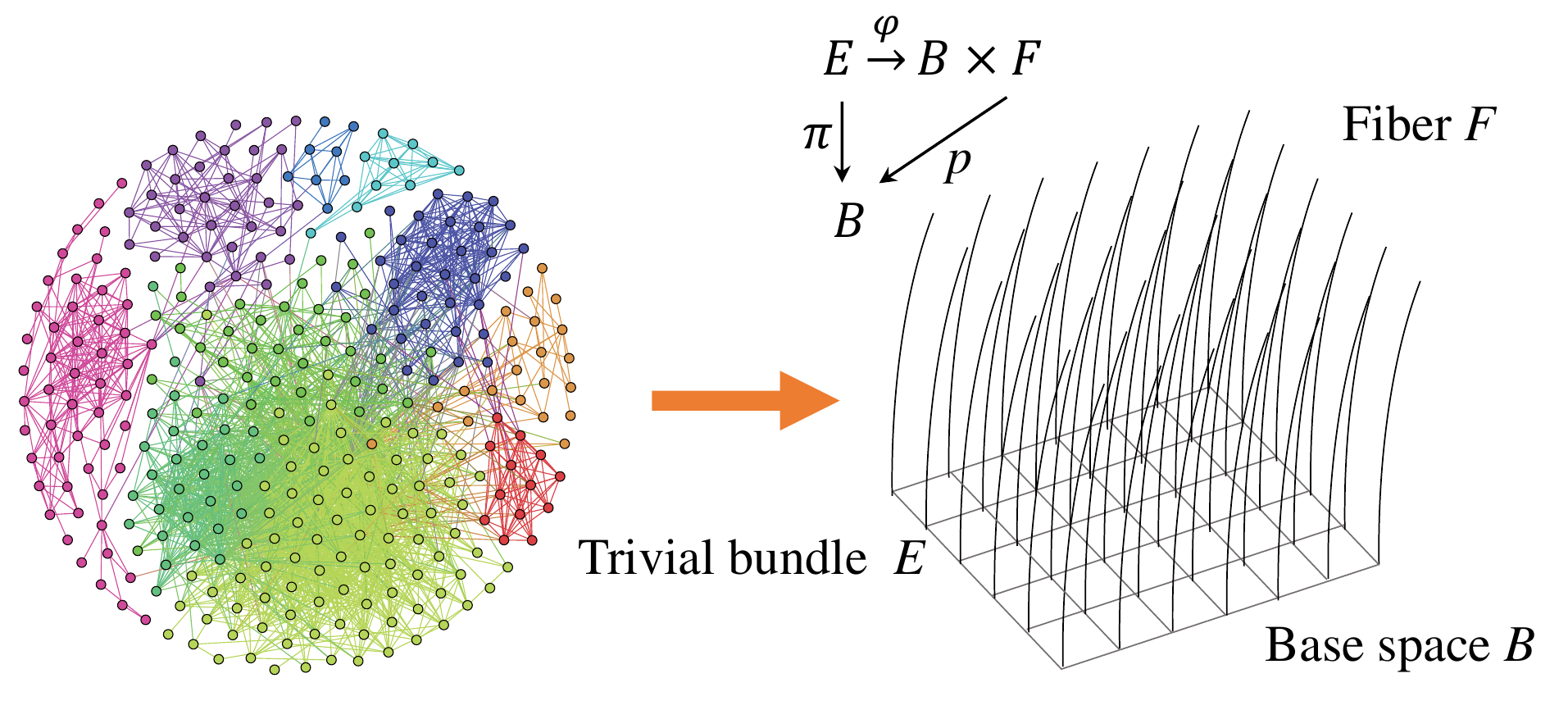} 
\caption{ The illustration of trivial bundle embeddings.   }
\label{fig1}
\end{figure*}

We consider graph representation learning on a   graph $G = (V, E,  X^0 )$, where $V$ is the node set, $E$ is the edge set, and    $X^0\in R^{|V|\times d}$ is the matrix of input node features.
The $i$-column ${(x^0_i)}$ of $X^0$ is the feature of   node $i\in V$ valued as a $d$-dimensional vector and the superscript$~^0$ indicates
the first layer.
When without input node features, ${(x^0_i)}$ is the one-hot vector with the value of the $i$-th entry being $1$ and all other entries being $0$.
Here, the goal in graph representation learning of TB-GCN  is to learn a mapping $f$ which embeds nodes to a trivial bundle $E=B\times F$   (Fig.\ref{fig1}). These embeddings should capture both structural and semantic information and can then be used as input for downstream tasks such as node classification and link prediction.

\subsection{Euclidean graph convolution networks}
For the implementation here, we only use  the Euclidean GCNs to achieve this aim.
Let $G = (V, E)$ be a graph with  node features $ X^{(0)} \in R^{N\times m}$. Let $(W^{(l)}, b^{(l)})$ be weights and bias parameters for layer $l$, and $\sigma(\cdot)$ be a non-linear activation function.
General GCN message passing rule at layer $l$ for node $i$ then is defined  as follows:
\begin{equation}
\label{linear_train}
X^{(l+1) }  =  \sigma\left( {\tilde{A}}  \left(X^{(l )}  W^{(l) }    +b^{(l)} \right)\right),
 \end{equation} where there exist different ways to compute the aggregation matrix $\tilde{A} $ based on network data.

\section {Literature review}
\subsection{Shallow Euclidean embeddings}
The early Euclidean embeddings for graphs are based on matrix-factorial approaches, directly inspired by the techniques of dimensional reduction.
The idea of these approaches is that considering a graph's adjacency matrix, for example,  the user-item interaction matrix of a bipartite graph,
it is decomposed into the product of two lower dimensionality matrices, then its nodes are represented as vectors in a lower-dimensional vector space.
 These approaches, such as singular value decomposition (SVD) and nonnegative matrix factorization (NMF)\citep{Lee1999}, are applied in  recommender systems\citep{Dhillon22005}, text clustering\citep{Gemulla2011}, etc.
 They belong to principal components analysis (PCA)\citep{Dunteman1989},
 however,
they do not consider the intrinsic geometry of the data.
  Laplacian eigenmaps are based on the assumption that
data are sampled from an intrinsic geometry lying in a high-dimensional space. They build a graph from $k$-nearest neighbors computed from the original data, which is a discrete approximation of its intrinsic geometry. Minimizing a cost function based on the graph ensures that points close to each other on the manifold are mapped close to each other in the low-dimensional space, preserving local distances\citep{Belkin2001}.


\subsection{Shallow non-Euclidean  embeddings}
Riemannian optimization has been used to learn embeddings to map graphs into Poincar\'e ball\citep{Nickel2017,Tifrea2019}  and Lorentz space\citep{Nickel2018}. These models outperform the Euclidean embedding methods on graph reconstruction and link prediction for scale-free networks, and have been applied to text analysis\citep{Tifrea2019,Dhingra2018}.
The maximum likelihood estimation  also has been used to embed graphs to  hyperbolic space\citep{Papadopoulos2015, Wang2016}. Coalescent embedding based on nonlinear  dimension reduction algorithms offers a fast  embedding in the hyperbolic space even for large graphs\citep{Muscoloni2017}.

\subsection{Deep Euclidean embeddings}
The methods reviewed above are shallow embedding approaches.They cannot share parameters for nodes, the number of parameters growing linearly with the number of nodes, fail to leverage node attributes, and are inherently transductive, not suitable for dynamic data.
 The Euclidean GCNs offer inductive approaches to learning embeddings from input graph structure as well as node features\citep{Xu2019}.
  These models can be designed as inductive ones\citep{Hamilton2017},
 as semi-supervised ones for classification task\citep{Kipf2017}, modified to use gated recurrent units\citep{Li2016} or attention mechanism\citep{Petar2018}, and applied to recommender systems\citep{Ying2018}.
   While resolving the disadvantages of shallow embeddings, these approaches lead to distortions when applied to scale-free networks.

\subsection{Deep non-Euclidean embeddings}
Based on hyperbolic neural networks\citep{Ganea2018} and hyperbolic attention networks\citep{Gulcehre2019}, GNNs have been extended to hyperbolic geometry, known as HGNNs, for graph classification tasks\citep{Liu2019}, link prediction, and node classification tasks\citep{Chami2019}.
Learning   embeddings to products of three kinds of spaces, namely spherical, hyperbolic, and Euclidean has been shown to outperform  the embeddings to single Euclidean or hyperbolic spaces\citep{Gu2019}.
Riemannian optimization is used to jointly learn the curvature and the embedding in the product space.
Although our work also discuss the embeddings to product  space, the loss function is different.
 For sequences of dimensions $s_1$,  $s_2$,..., $s_m$, $h_1$,  $h_2$,..., $h_n$, and $e$, the embedding space is
$P = S^{s_1} \times S^{s_2} \times\cdots \times S^{s_m}\times H^{h_1} \times H^{h_2} \times\cdots\times H^{h_n}\times  E^e$;
the loss function in Ref.\citep{Gu2019} is
$L(x) =  \sum_{
1\leq i\leq j\leq n}| d_P(x_i,x_j)^2/  |
 d_G(X_i,X_j)^2 -1 |$, where $d_P (p, q) = \sum^{s_m}_{i=1}  d_{S_i}(p, q) +  \sum^{h_n}_{j=1}  d_{H_j}(p, q) + d_E(p, q)$, and $d_G(X_i, X_j)$ is the  given graph distance.

\section{TB-GCN architecture}

\begin{figure*}[t]
\centering
\includegraphics[height=5.5    in,width=3.8  in,angle=0]{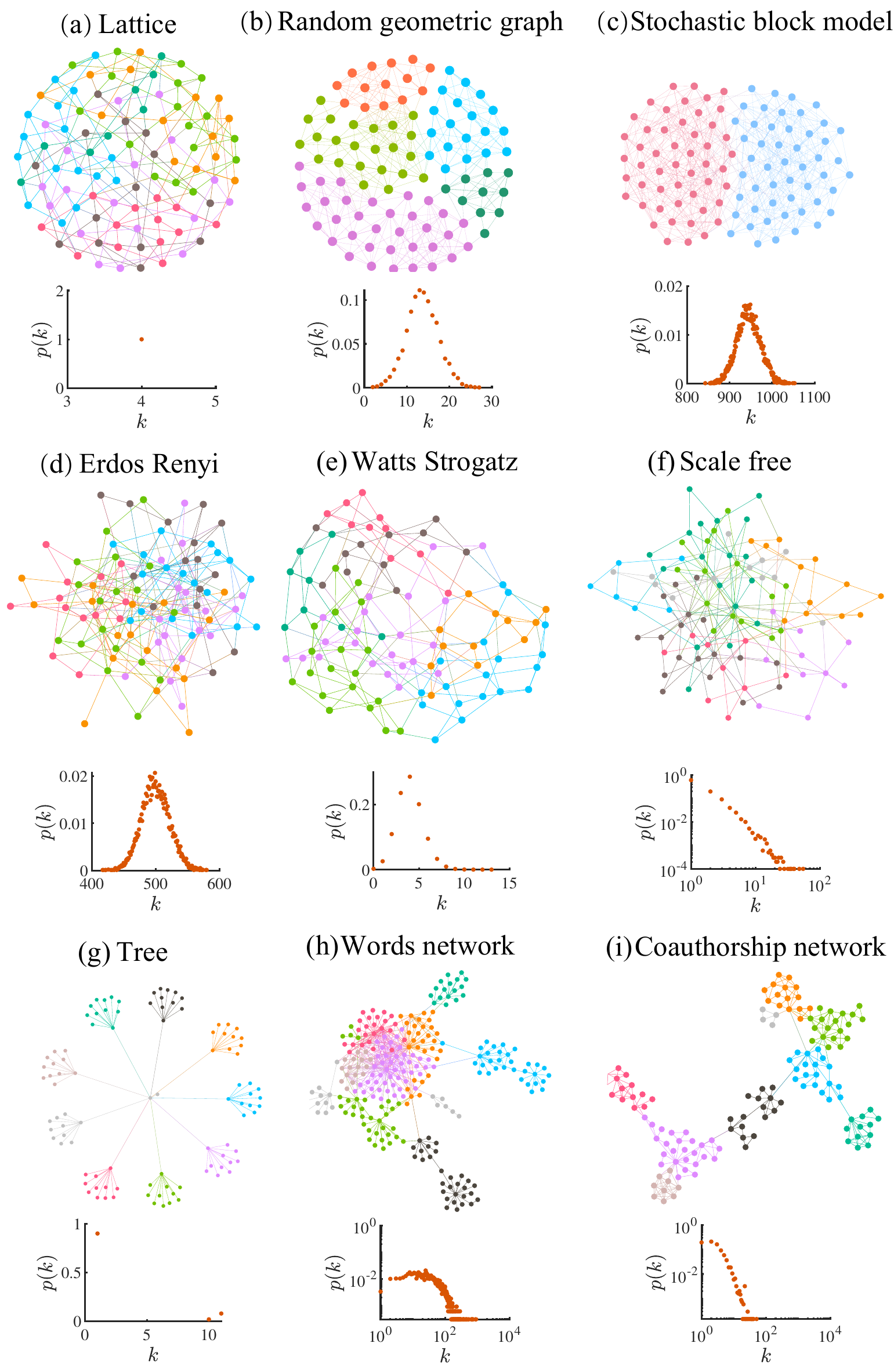} 
\caption{ Typical real-word and synthetic networks and their  degree distribution.
The degree distributions of the  networks from
   different  types     can belong to the same class. }
\label{fig2}
\end{figure*}

We propose TB-GCN to embed networks into trivial bundles,   representing the assortativity and disassortativity in the preference of choosing which nodes to connect.
This is a perspective different from HGCNs, which are proposed based on the assumption that the underlying geometry of scale-free networks is hyperbolic. That is, the rationality of HGCNs is based on the scale-free feature of networks.
Given a network with a fixed degree distribution,  we can tune its assortativity coefficient by rewiring a fraction of connections\citep{Newman2003}.
Fig.\ref{fig2} shows   the assortativity  coefficient is free of
the feature of the degree distribution, such as scale-free, Poisson, and their mixture.
Therefore,  discussing which space is suitable to embed a network with a given assortativity coefficient has its own meaning.

\begin{figure}[t]
\centering
\includegraphics[height=2.3   in,width=4  in,angle=0]{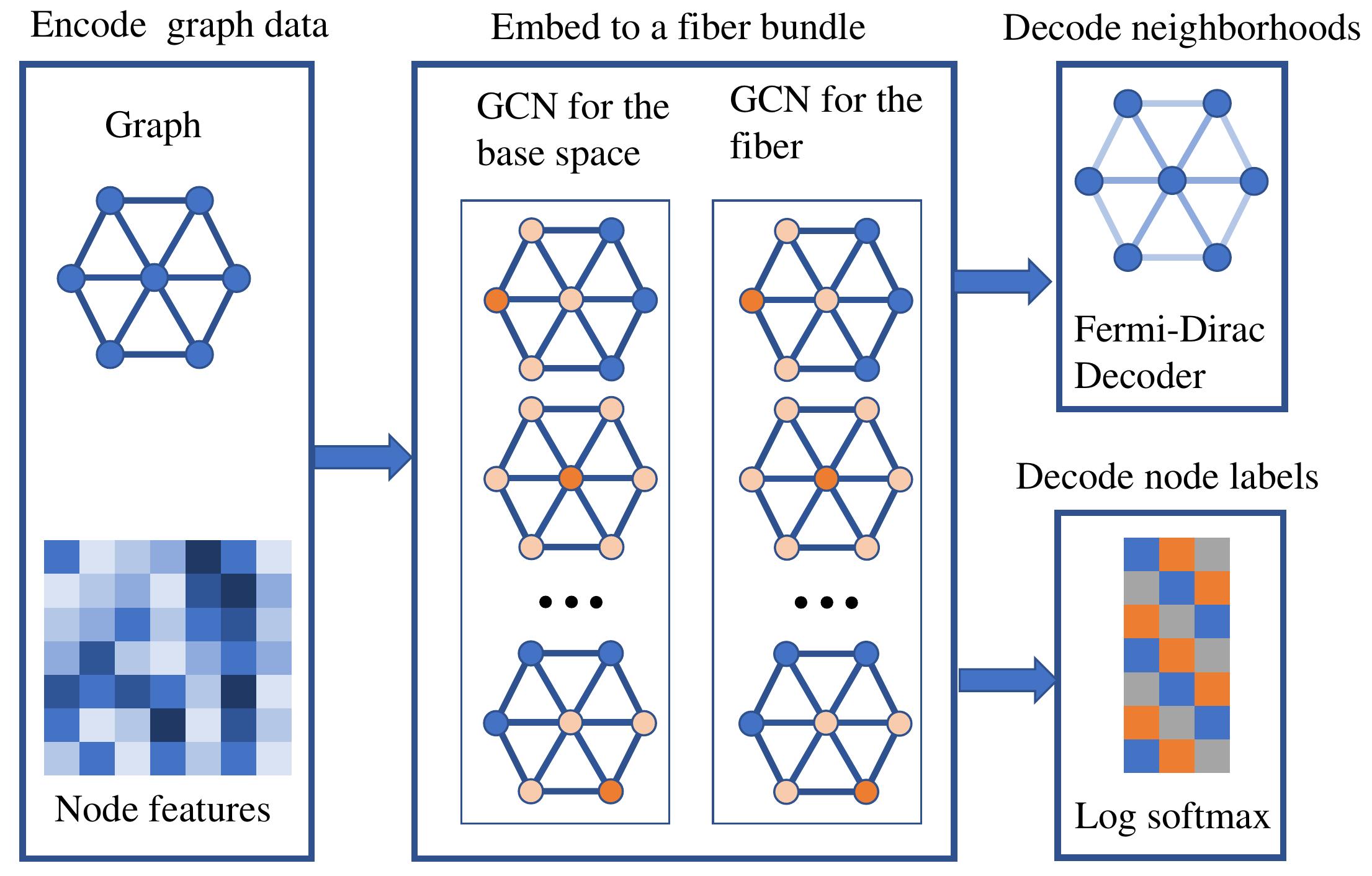} 
\caption{The architecture of TB-GCN. The originality is the  decoders, which   represent the assortativity and disassortativity   in nodes' preference of choosing which nodes to connect.   }
\label{fig3}
\end{figure}

  We assume that    both assortativity and disassortativity exist
 in the preference of choosing which nodes to connect. We learn embeddings of network data such that their coordinates of base space can be used to decode the assortativity and their coordinates of fiber used to decode the disassortativity.
  There exist multiple choices of base space and fiber, for example the  Poincar\'e ball for base space and the Euclidean for fiber.
  In the following, we will consider a simple case, namely, both are the Euclidean space, as it is enough to represent the assortativity and disassortativity.
The architecture of
TB-GCN  is described within the encoder-decoder framework (Fig.~\ref{fig3} ), The originality of TB-GCN  is its   decoders, which represent  the assortativity  and disassortativity. The model is trained by minimizing the loss function of a given task.

\subsection{Encoders}

 We describe our main embedding space: $E=B\times F$, where $B=\{x_1,...,x_m\}$
and $F=\{y_1,...,y_n\}$.
Given a graph
$G = (V, E)$ and input Euclidean features $\{X^{(0)} \in R^{N\times m}  ,Y^{(0)}   \in R^{N\times n}\}$.
Let $\tilde{A}  = D^{-\frac{1}{2}}  (A + I)D^{-\frac{1}{2}}  $, where $I$ is the identity matrix,  $A$ the adjacency matrix, and  $D=( \sum_j(A_{ij} + I_{ij}))$.
TB-GCN   stacks multiple Euclidean  graph convolution layers for the base space and fiber with the following layer-wise propagation rule:
    \begin{equation}
\left\{\begin{aligned}
\label{propagation}
X^{(l+1) } &=  \sigma\left( {\tilde{A}}  \left(X^{(l )}  W^{(l) }_B   +b^{(l)}_B\right)\right),\\
Y^{(l+1) } &=  \sigma\left(  {\tilde{A}}  \left(Y^{(l) } W^{(l) }_F  +b^{(l)}_F\right)\right) .
\end{aligned}\right.
\end{equation}
 Here   $W^{(l)}_{B,F}$ and $b^{(l)}_{B,F}$  are the layer-specific trainable weight matrices and biases for the base space $B$ and $F$ respectively; $\sigma(\cdot)$
denotes an activation function, such as the $\mathrm{ReLU}(\cdot) = \max(0,\cdot )$; $X^{(l)}\in R^{N\times m}$ and
 $Y^{(l)}\in R^{N\times n}$
 are the matrices of activations in the $l$-th layer; $X^{(0)}$ and $Y^{(0)}$ are input.
Note that
our model is   a multi-view model if $X^{(0)}\neq Y^{(0)}$, and not  if $X^{(0)}=Y^{(0)}$.

\subsection{Decoders}
If the objective is the probability of existing an edge between two given nodes,
We use the Fermi-Dirac decoder  to decoder the probability
scores for edge $l=(p,q)$:
    \begin{equation}
\label{probability}  P(l\in E ) = \left ( {\mathrm{e}^{( {  d( p , q ) -  r })/{  t}} + 1 }\right )^{-1},\end{equation}
where $d(p,q)$ is a value positively correlated to  $ P(l\in E )$, $r$ and $t$ are hyper-parameters.
One originality of our model is defining
\begin{equation}
\label{dist} d( p , q) =\left (\sum^m_{i=1}(x_i(p)-x_i(q))^2 \right )   \left (\sum^n_{j=1}(y_j(p)+y_j(q))^2 \right ).
\end{equation}
The first  factor   represents the assortativity,  and the second  disassortativity, and thus $d( p , q)$ both. That is, the role of base space embeddings  is used to decode the assortativity and the role of fiber embeddings is to  the disassortativity.
  Note that
  Eq.\ref{dist} does not need $m=n$.

  Different from the product space embeddings\citep{Gu2019}, the decoder here not only considers the assortativity factor (the smaller the distance between two nodes, the more likely they connect)  but also considers the disassortative factor  (the larger the distance between two nodes, the more likely they connect).
   The roles of the first and
second terms in Eq.\ref{dist} are equal. Which term or both being activated depends on the assortativity, disassortativity, or their
mixture of data.

If the objective is a vector that is used to represent the label or type of a node, we decode the vector by subtracting, dividing, and multiplying each element of the last layer of base space $X^{(L) }$  by the corresponding element of the last layer of fiber $Y^{(L) }$ respectively.
 The subtracting and dividing operations represent the antagonism of the base space embeddings and fiber embeddings, meanwhile, the multiplying operation represents their complementarity. Then, in those situations,  the dimension of fiber should be equal to that of base space.

\subsection{Two Tasks}

\textbf{Link prediction.}
For LP tasks, We use negative sampling to  train TB-GCN  by minimizing the cross-entropy loss.
Let $E^+$ and $E^-$ be the positive and negative training  edge sets.
The loss is
\begin{equation}
\label{loss} L_{\mathrm{LP}}=-\sum_{l\in E^+} \frac{ \log P(l\in E^+ )}{|E^+|}-\sum_{l\in E^-} \frac{\log(1-P(l\in E^-  ))}{|E^-|} .
\end{equation}

\textbf{Node classification.}
For NC tasks, we have three ways to decode the embeddings as vectors, namely subtracting, dividing, and multiplying, and we denote the corresponding models as TB-GCN-SUB, -DIV, and -MUL respectively. Therefore, the dimension of fiber should be equal to that of base space for NC tasks, namely $m=n$.

 We perform the Euclidean multinomial logistic regression on outputs, and then train TB-GCN  by minimizing the negative log likelihood loss    \begin{equation}
\label{losscn} L_{\mathrm{CN}}= -\sum_{i\in N^\mathrm{T}} \frac{ \log P_{c(i)}(i)}{| N^\mathrm{T}|},
\end{equation} where $P_{c(i)}(i)$ is the  probability that node $i$ is correctly classified into its class $c(i)$.
    We also add a link
prediction regularization objective in node classification tasks, to encourage embeddings at the last layer to preserve the
graph structure. In this situation, the loss is \begin{equation}\label{lossCNLP} L_{\mathrm{CNLP}}= \gamma L_{\mathrm{CN}}+(1-\gamma) L_{\mathrm{LP}}, \end{equation}
where $\gamma\in[0,1]$  weights the role of preserving    the
graph structure in NC tasks.

\textbf{Computational complexity.}
The number of multiplications in Eq.~(\ref{propagation}) is $N^2(m+n+1)+N(m^2+n^2)$, and thus  the number of multiplications
of   propagating $K$ layers is $KN^2(m+n+1)+KN(m^2+n^2)$.
 Therefore, the complexity of TB-GCNs is on
par with  that of GCNs. It can be scaled to large graphs as GCNs do.

\section{Experimental setup}
\subsection{Datasets } We use a variety of   datasets that we detail in Tables \ref{tab1} and \ref{tab2}.
We compute their assortativity coefficient, an index that measures the degree of assortativity a graph is.
The lower, the more disassortativity is the network.
The random geometric graph and coauthorship networks have a positive assortativity coefficient, and trees negative.

\begin{table}[t]
\setlength\tabcolsep{2.2pt}
\centering
\begin{tabular}{l|c|c|c|c|c }
\hline
Network &Nodes &Edges& Classes &Features & Assortativity  \\ \hline
CORA &2708 &5429 &7& 1433   &  -0.0659   \\
PUBMED &19717& 44327& 3& 500 &  -0.0436  \\        
AIRPORT &3188 &18631 &4 &4 &-0.0157   \\
DISEASE-CN &1044& 1043& 2& 1000  & -0.5441 \\\hline
\end{tabular}
\caption{Statistics of the networks used  for  node classification tasks.}
\label{tab1}
\end{table}

\begin{table}[t]
\setlength\tabcolsep{2.2pt}
\centering
\begin{tabular}{l|c|c|c}
\hline
Name &Nodes &Edges&  Assortativity  \\ \hline
WORD-TREE &4679 &128974 &   -0.1128   \\
CA-HEP  &11204& 117619& 0.6295  \\
CA-INF  &13102 &19239 &0.1627 \\
WIKI-VOTE&7066&100736&-0.0833 \\
DISEASE-LP &2665& 2664& -0.6135   \\
WS&1600&3200&0.0116 \\  
PA& 1000 & 7000 &-0.0219 \\                         
BSM&1000&52411&-0.0100\\
RGG&1000&  14233& 0.5423\\                     
TREE&1000&999&-0.8223   \\                         
TREE+LATTICE& 1000& 2355&-0.0441  \\              
TREE+RGG&1000& 2407& -0.0312  \\ 
\hline
\end{tabular}
\caption{Statistics  of the networks used for link prediction tasks.}
\label{tab2}
\end{table}

1. \textbf{Citation networks.} CORA\citep{Sen2008} and PUBMED\citep{Namata2012} are standard benchmarks describing citation networks, where nodes represent papers, edges represent citations between them, and node labels are academic areas.

2.  \textbf{Coauthorship networks.}  CA-HEP is  standard benchmark describing collaboration networks covering papers in the period from Jan 1993  to Apr  2003\citep{Leskovec-TKDD2007}, where
nodes represent authors, edges represent coauthorship between them.
CA-INF downloaded from   Web of Science\footnote[2]{http://www.webofknowledge.com}, which
covers authors' collaborations in the papers  of three accredited
journals of informetrics, namely the Journal of the Association for Information Science and Technology,
Journal of Informetrics, and Scientometrics during Jan  2006-- Dec 2002. If the paper is co-authored by $k$ authors this generates a completely connected subgraph on $k$ nodes.

3. \textbf{Tree-like networks.} Disease-CN and -LP are  tree-like networks   generated  by the SIR disease spreading model\citep{Anderson1992},
where the label of a node is whether
the node was infected or not and node features indicate the
susceptibility to the disease.
WORD-TREE is a word-attention network containing the word branches of the PNAS papers published in 2015, which are generated by the model in Ref.~\citep{Xie2021}.

4. \textbf{Flight  networks.} AIRPORT is a transductive dataset where nodes represent airports and edges represent the airline
routes as from Open Flights.org.  The network has geographic information (longitude, latitude, and altitude), and the GDP of the country
where the airport belongs to. Node labels are the populations of airports' country\citep{Chami2019}.

5. \textbf{Vote network.}
WIKI-VOTE contains all the Wikipedia voting data from the inception of Wikipedia till 2008.01\citep{Leskovec-Huttenlocher2010}. Nodes in the network represent Wikipedia users and a directed edge from node $i$ to node $j$ represents that user $i$ voted on user $j$.

6.  \textbf{Synthetic  networks.}  WS is generated by the Watts-Strogatz model, where the dimension of the lattice is  2, the size of the lattice along all dimensions is 10,
the value giving the distance within which two nodes will be connected is 1, and the rewiring probability is 0.1.
 SCALE-FREE is a scale-free network, and the exponent of its out-degree distribution is 2.1. BSM is generated by
the stochastic block model, where
the number of blocks is 2, the number of  nodes  in each block is 500,
the	matrix giving the connection probabilities for different blocks is $[0.2,0.01]$. RGG is a random geometric graph in a disc with radius 1,
and two nodes will connect if their distance   is less than 0.03.
TREE is a network in which almost all nodes have the same number of children 10. TREE+LATTICE is the TREE with edges generated by lattice model with dimensional list $[10,10]$.
TREE+RGG is the combination of TREE and RGG.


\subsection{Baselines }
Since the deep methods have been shown to outperform shallow methods, we only consider six state-of-the-art deep methods:  Graph attention networks (GAT)\citep{Petar2018}, Graph convolution networks (GCN)\citep{Kipf2017},   Hyperboloid and Poincar\'e embeddings  \citep{Nickel2017}  with hyperbolic neural networks \citep{Ganea2018}    and hyperbolic graph convolution networks\citep{Chami2019}  respectively (HNN-HYP, HNN-POI, HGCN-HYP, HGCH-POI). We conjecture that TB-GCN  will outperform  on
the networks with both of the assortative and disassortative existing in nodes' preference of choosing which nodes to connect.

The propose of our model is to show that compared with Euclidean and hyperbolic spaces, trivial bundles would be suitable for GCN embedding the complex networks with both of the assortative and disassortative existing in nodes' preference of choosing which nodes to connect. We mainly compared it with GCNs.
Meanwhile, assertion error happens when running GATs on some datasets, e.g., CORA. Therefore,   some  state-of-the-art of hyperbolic GATs \citep{zhu2020bilinear,zhang2021hyperbolic} are not chosen as baselines.

\subsection{Implementation Details}

The hyper-parameters of $t$ and $r$ in Eq.~(\ref{probability}), the initial learning rate, weight
decay, DropConnect\citep{Wan2013}, the number of layers, and activation functions  are the same as HGCN\citep{Chami2019}.
 The hyper-parameter $\gamma\in \{0.0,0.25,0.50,0.75,1.0\}$ in the loss (\ref{lossCNLP}) is
used only for the NC task.
The     dimension  of embedding space  is 128 here,  but 16 in Ref.~\citep{Chami2019}).

We use early stopping based on validation set performance with the patience of 1,000 epochs and 15,000 the maximum number of epochs to train. We measure performance on the final test set. For fairness, we   control the   dimensions to be the same (128)
for all methods,    the dimension  of fiber and that of base space to be 64. We use a fixed random seed set 1234.
We optimize the considered    TB-GCN (note that the base space and fiber   are Euclidean), GCN, GAT with Adam Optimizer\citep{Kingma2015},
and others with
RiemannianSGD Optimizer\citep{Bonnabel2013,Zhang2016}.
We open-source our implementation of TB-GCN at Github, which contains the initial settings of hyperparameters of   TB-GCN and baselines.

\begin{figure*}[t]
\centering
\includegraphics[height=2.3    in,width=4.5   in,angle=0]{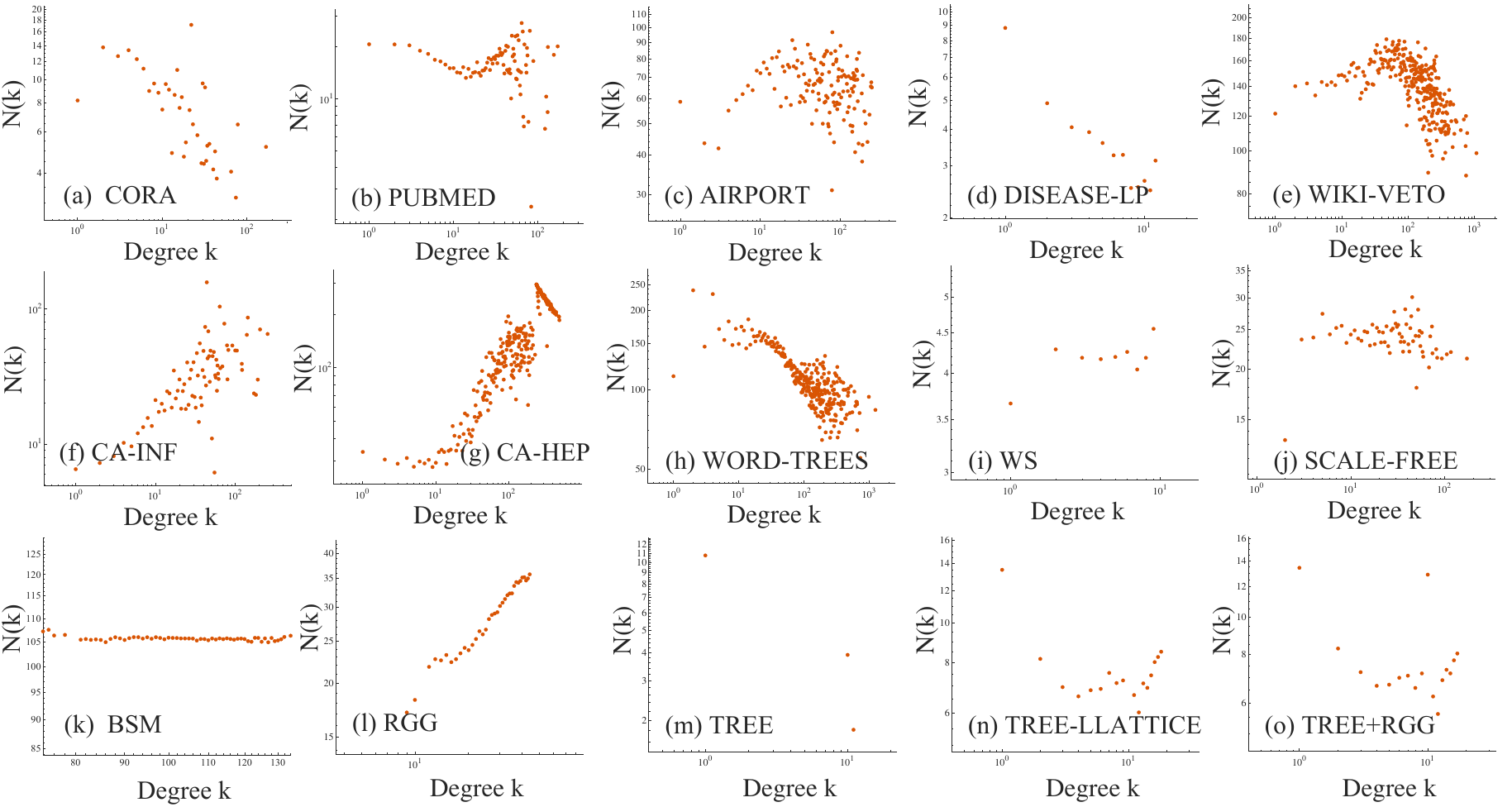} 
\caption{ The average degree $N(k)$ of  $k$-degree nodes' neighbors. The panels show $N(k)$ for the networks in Table~\ref{tab2}.}
\label{fig4}
\end{figure*}

\subsection{Evaluation Metric }
We evaluate our method on a variety of networks, on both  NC and  LP  tasks.
In LP tasks, we randomly split edges into 85/5/10\% for training, validation, and test sets, the same as those in Ref.\citep{Ioffe2015}, and use the node features if exist. For   NC tasks, we use 70/15/15\% splits for AIRPORT, 30/10/60\% splits for DISEASE, and
standard splits in Refs.\citep{Kipf2017,Yang2016} with 20 train examples per class for CORA and PUBMED.
We use the node features if exist.
We evaluate link prediction by measuring the area under the
ROC curve (AUC) and average precision (AP) on the test set and evaluate node classification by measuring F1 score.

\subsection{Overall Performance}

Table~\ref{F1-TB-GCN} reports the performance of TB-GCN with different decoders (-SUB, -DIV, -MUL) and  different values of the parameter  $\gamma$ in the loss~(\ref{lossCNLP}) on NC tasks. The cases with $0<\gamma<1$ outperform those with $\gamma=0$, which means both node features and network topology contribute to NC tasks.
TB-GCN-SUB achieves the best performance on two networks, and others on one, which means all of the  decoders have their advantages.

\begin{table*}[!t]
\setlength\tabcolsep{2.2pt}
\centering
\begin{tabular}{l|c|cccc}
\hline
~ & $\gamma$ & CORA & PUBMED & AIRPORT & DISEASE\textunderscore CN \\
\hline
\multirow{5}*{TB-GCN-SUB} & 0 & 0.7880 & 0.7590 & 0.9294  & 0.8734\\
~ & 0.25 & 0.7980 & \textbf{0.8030} & \textbf{0.9370 }& 0.8634\\
~ & 0.50 & 0.8040 & 0.7990 & 0.9313 & 0.7814\\
~ & 0.75 & 0.8050 & \textbf{0.8030} & 0.9084 & 0.7005\\
~ & 1.00 & 0.1320 & 0.3890 & 0.2729 & 0.8053\\
\hline
\multirow{5}*{TB-GCN-DIV} & 0 & 0.6990 & 0.7080 & 0.8378  & 0.7368\\
~ & 0.25 & 0.7440 & 0.7480 & 0.7805 & \textbf{0.9091}\\
~ & 0.50 & 0.7460 & 0.7620 & 0.7882 & 0.8974\\
~ & 0.75 & 0.7480 & 0.7670 & 0.7538 & 0.8161\\
~ & 1.00 & 0.2960 & 0.5650 & 0.4351 & 0.5141\\
\hline

\multirow{5}*{TB-GCN-MUL} & 0 & 0.7960 & 0.7550 & 0.9160  & 0.8734\\
~ & 0.25 & 0.8060 & 0.7850 & 0.9179 & 0.8596\\
~ & 0.50 & \textbf{0.8220} & 0.7780 & 0.9160 & 0.8496\\
~ & 0.75 & 0.8190 & 0.7850 & 0.9160 & 0.8235\\
~ & 1.00 & 0.2450 & 0.3890 & 0.1756 & 0.7401\\
\hline
\end{tabular}
\caption{F1 scores of TB-GCN   on  node classification tasks for the networks in Table~\ref{tab1}. We report the   performances of   TB-GCN with different values of the parameter $\gamma$ in the loss~(\ref{lossCNLP}) and different decoders -SUB, -DIV, -MUL.}
\label{F1-TB-GCN}
\end{table*}

Table~\ref{F1} reports the performance of baseline methods on NC tasks in comparison to the best performance of TB-GCN.
It shows that TB-GCN achieves error reduction compared to the best baselines for the four networks considered in Ref.~\citep{Chami2019}, especially on DISEASE-CN (error reduction 6\%). The   variations between the performance of baseline approaches here and the performance in the results are computed by the package "gcn-master" that is downloaded from Github.  Ref.~\citep{Chami2019} due to the difference in the dimensions of embedding space (128 here  and 16 in Ref.~\citep{Chami2019}).

\begin{table*}[t]
\setlength\tabcolsep{2.2pt}
\centering
\resizebox{1.00\columnwidth}{!}{
\begin{tabular}{l|c|c|c|c|c|c|c|c}
\hline
  &TB-GCN   &HGCN-HYP& HGCN-POI &HNN-HYP& HNN-POI &GCN&GAT & ERR-RED\\
\hline
CORA & \textbf{0.8220} & 0.7740 & 0.7930 & 0.5910 & 0.6070 & 0.7990 & 0.7540  &0.0230\\
PUBMED & \textbf{0.8030} & 0.7830 & 0.7770 & 0.7390 & 0.7350 & 0.7570 & 0.7370& 0.0200\\
AIRPORT & \textbf{0.9370} & 0.8511 & 0.8359 & 0.7939 & 0.7767 & 0.9218 & 0.8153& 0.0152\\
DISEASE-CN & \textbf{0.9091} & 0.7228 & 0.7714 & 0.6667 & 0.6667 & 0.8444 & 0.8444 &0.0647\\\hline
\end{tabular}
}
\caption{ F1 scores of  baselines on node classification tasks for the networks in Table~\ref{tab1}. We report the best performance of TB-GCN  compared with baselines. }
\label{F1}
\end{table*}

Table~\ref{AUC}   reports the performance of TB-GCN  on LP tasks in comparison to the performances of baseline methods.
It shows that TB-GCN achieves error reduction compared to the best deep baselines for   eight networks in Table~\ref{tab1}.
The performances of baseline methods on the first four networks here
slightly vary from that in Ref.~\citep{Chami2019}, which are also due to the difference in hyperparameter settings.
HNN-POI outperforms TB-GCN on DISEASE-LP, suggesting that hyperbolic geometry is more suitable for the SIR disease spreading model.
TB-GCN outperforms baseline methods on the real-world network WORD-TREES, the synthetic networks TREE, and TREE+RGG, suggesting that TB-GCN is also suitable for the networks with tree-like structures.

\begin{table*}[t]
\centering
\resizebox{1.0\columnwidth}{!}{
\begin{tabular}{l|c|c|c|c|c|c|c|c|c}
\hline
Data & ~ & TB-GCN & HGCN-HYP & HGCN-POI & HNN-HYP & HNN-POI & GCN & GAT & ERR RED\\
\hline
\multirow{2}*{CORA} &  {AUC} & \textbf{0.9409} & 0.9196 & 0.9167 & 0.9021 & \textbf{0.9409} & \textbf{0.9409} & 0.9400 &0.000\\  
~ &  {AP} &\textbf{ 0.9424} & 0.9249 & 0.9286 & 0.9141 & \textbf{0.9424} & \textbf{0.9424} & 0.9382  &0.000\\
\hline
\multirow{2}*{PUBMED} &   {AUC} & 0.9588 & 0.9414 & 0.9442 & 0.9619 &\textbf{ 0.9650} & 0.9384  & 0.9439&-0.0062\\  
~ &  {AP} & 0.9586 & 0.9451 & 0.9491 & 0.9569 &\textbf{ 0.9607 }& 0.9349 & 0.9415&-0.0021\\
\hline
\multirow{2}*{AIRPORT} & {AUC} & \textbf{0.9576} & 0.9332 & 0.9394 & 0.9471 & 0.9316 & 0.9536 & 0.9471&0.004\\    
~ &  {AP} & \textbf{0.9517 }& 0.9447 & 0.9415 & 0.9426 & 0.9389 & 0.9473 & 0.9426&0.0044\\
\hline
\multirow{2}*{DISEASE-LP}&  {AUC} & 0.6339 & 0.6159 & 0.6561 & 0.7448 & \textbf{0.8033} & 0.6181 & 0.6308&-0.1694\\    
~ &  {AP} & 0.6025 & 0.5998 & 0.6325 & 0.7290 &\textbf{ 0.7705} & 0.6050 & 0.6162&-0.1680\\
\hline
\multirow{2}*{WIKI-VETO} & {AUC} &\textbf{ 0.9701 }& 0.9525 & 0.9499 & 0.9541 & 0.9579 & 0.9431 & 0.9413 &0.0122\\  
~ &  {AP} & \textbf{0.9653} & 0.9542 & 0.9504 & 0.9556 & 0.9572 & 0.9219 & 0.9223& 0.0081\\
\hline
\multirow{2}*{CA-INF} &  {AUC} &\textbf{ 0.9481 }& 0.8922 & 0.9153 & 0.8930 & 0.9056 & 0.9357 & 0.9408 &0.0073 \\ 
~ &  {AP} & \textbf{0.9648} & 0.9386 & 0.9503 & 0.9417 & 0.9469 & 0.9565 & 0.9531&0.0117 \\
\hline
\multirow{2}*{CA-HEP} & {AUC} & \textbf{0.9818} & 0.9757 & 0.9761 & 0.9755 & 0.9782 & 0.9700 &  0.9742& 0.0036\\  
~ & {AP} & \textbf{0.9860} & 0.9839 & 0.9841 & 0.9832 & 0.9849 & 0.9703 & 0.9744 & 0.0011\\
\hline
\multirow{2}*{WORD-TREES} &  {AUC} &\textbf{ 0.8690} & 0.8668 & 0.8671 & 0.8628 & 0.8652 & 0.8310 & 0.8520&0.0019\\ 
~ &  {AP} & \textbf{0.8498} & 0.8416 & 0.8427 & 0.8371 & 0.8394 & 0.8097 & 0.8323&0.0071\\
\hline
\multirow{2}*{WS} &  {AUC} & 0.7524&\textbf{ 0.7659}  & 0.7075 & 0.7080 & 0.7023 & 0.7356 & 0.7371 &-0.0135\\  
~ &   {AP} & \textbf{0.7889 }& 0.7073 & 0.7685 & 0.7613 & 0.7603 & 0.7646 & 0.7748&0.0141\\
\hline
\multirow{2}*{SCALE-FREE} &  {AUC}  & 0.6470 & \textbf{0.6669} & 0.6643 & 0.6500 & 0.6589 & 0.6389 & 0.6371  &-0.0026\\   
~ &  {AP} & 0.6639 &\textbf{ 0.6807 }& 0.6791 & 0.6641 & 0.6739 & 0.6390 & 0.6465&0.0016\\
\hline
\multirow{2}*{BSM} &  {AUC} & 0.7602 & 0.7613 & \textbf{0.7636 }& 0.7588 & 0.7614 & 0.7618 & 0.7582&-0.0034 \\  
~ & {AP} & 0.6781 & 0.6810 & \textbf{0.6859} & 0.6752 & 0.6817 & 0.6860 & 0.6738&0.0078\\
\hline
\multirow{2}*{RGG} & {AUC} & 0.9966 &\textbf{ 0.9969 }& 0.9968 & 0.9965 & 0.9964 & 0.9939 & 0.9943&-0.0003 \\   
~ & {AP} & 0.9888 & \textbf{0.9893} & 0.9891 & 0.9888 & 0.9887 & 0.9850 & 0.9858 &-0.0005   \\
\hline
\multirow{2}*{TREE} &  {AUC} & \textbf{0.8045 }& 0.6673 & 0.6650 & 0.6556 & 0.6540 & 0.6389 & 0.6371&0.1372\\   
~ &   {AP} & \textbf{0.7075} & 0.6800 & 0.6788 & 0.6745 & 0.6700 & 0.6390 & 0.6465&0.0275\\
\hline
\multirow{2}*{TREE+LATTICE} &   {AUC} &  \textbf{0.9528} & 0.9141 & 0.9136 & 0.9196 & 0.9208 & 0.9044 & 0.9008 &0.032\\ 
~ &   {AP} &  \textbf{0.9574} & 0.9366 & 0.9361 & 0.9388 & 0.9398 & 0.9189 & 0.9153 &0.0176\\
\hline
\multirow{2}*{TREE+RGG} &  {AUC} &\textbf{ 0.9172} & 0.8979 & 0.8989 & 0.9027 & 0.9098 & 0.8731 & 0.9040&0.0074\\ 
~ &  {AP} & \textbf{0.9186 }& 0.9078 & 0.9093 & 0.9136 & 0.9164 & 0.8730 & 0.8931&0.0022\\
\hline
\end{tabular}
}
\caption{  AUC and AP of   TB-GCN  and baselines on   link prediction tasks for the networks in Table~\ref{tab2}.  }
\label{AUC}
\end{table*}

Table~\ref{AUC} reports the best baselines to outperform TB-GCN on DISEASE-LP, BSM, RGG, WS, and SCALE-FREE.
We analyze the relationship between a node's degree and its neighbors' degree. We consider denoting the average degree
of $k$-degree nodes' neighbors by $N(k)$ (Fig~\ref{fig4}), and find that the    these networks' $N(K) $ are close to  linear functions.
Therefore, we conjecture that HGCN works better on networks with non-linear $N(K)$. For these networks,  the connecting behaviors of some nodes are featured by assortativity and some by disassortativity, for example, scientific coauthorship networks. We refer the readers to Ref.~\citep{XieOu2018} for a thorough discussion of the transition phenomena of  $N(K)$ for coauthorship networks.

\section{Conclusion}
We introduced a novel architecture TB-GCN  that learns the trivial bundle embeddings for networks with or without node features using graph convolutional networks.
TB-GCN represents both the assortativity and disassortativity in nodes' preference of choosing which nodes to connect, a new perspective of finding a proper underlying geometry for networks.
It achieves state-of-the-art  or comparable performances  on LP and NC tasks    in comparison to   the Euclidean and hyperbolic GCNs,
suggesting its ability to learn geometric representations for networks, especially for the networks with some nodes' connecting behaviors featured by assortativity and some by disassortativity.
We expect that non-trivial bundle embeddings can further increase the quality of bundle embeddings.

\bibliography{References}


\end{document}